\documentclass{article}

\PassOptionsToPackage{sort,numbers, compress}{natbib}
\usepackage[preprint]{neurips_2026}

\usepackage[utf8]{inputenc} 
\usepackage[T1]{fontenc}    
\usepackage{hyperref}       
\usepackage{url}            
\usepackage{booktabs}       
\usepackage{amsfonts}       
\usepackage{amsmath}       
\usepackage{amssymb}   
\usepackage{pifont}    
\usepackage{tabularx}
\usepackage{multirow}

\usepackage{bbm}       
\usepackage{nicefrac}       
\usepackage{microtype}      
\usepackage{xcolor}         
\usepackage{graphicx}       
\usepackage{algorithm}
\usepackage{algpseudocode}
\usepackage[most]{tcolorbox}

\newtcolorbox[auto counter, number within=section]{promptbox}[2][]{
  colback=black!4, colframe=black!55, boxrule=0.5pt,
  fonttitle=\bfseries\small, fontupper=\small, breakable,
  title={Box~\thetcbcounter: #2}, label={#1}}

\title{
MLIP Detective: Active Failure Mode Discovery Beyond Benchmark Scores for Machine-Learning Interatomic Potentials
}

\author{%
  Ryuhei Okuno \\
  Preferred Networks\\
  \texttt{ok79ryuhei@preferred.jp} \\
  \And
  Nontawat Charoenphakdee \\
  Preferred Networks \\
  \texttt{nontawat@preferred.jp} \\
  \And
  Kaoru Hisama\\
  Preferred Networks \\
  \texttt{hisama@preferred.jp} \\
  \And
  Yuta Tsuboi \\
  Preferred Networks \\
  \texttt{tsuboi@preferred.jp} \\  
}

\hypersetup{hidelinks}

\begin{document}

\maketitle

\begin{abstract}
Universal machine-learning interatomic potentials (u-MLIPs) aim to generalize across diverse configurations.
Benchmarks enable reproducible evaluation but may not expose failures outside their predefined scope.
Here, we show that physics-informed search can complement benchmark-based evaluation by uncovering hidden failure modes.
We introduce \textbf{MLIP Detective}, an agentic framework for active failure mode discovery.
Starting from benchmark evidence, MLIP Detective generates falsifiable, physics-informed failure hypotheses, screens them with inexpensive simulations, and escalates only the most suspicious cases to human experts together with proposed verification protocols.
Without issue-specific prompting, MLIP Detective identified and characterized a systematic anomaly in MACE-MPA-0: the model predicted some relaxed adsorbate--surface systems involving O- or F-containing adsorbates to be higher in energy than their corresponding separated fragments.
Using cross-model comparisons, MLIP Detective further inferred a likely training-data origin for the anomaly, consistent with recent reports.
\end{abstract}

\section{Introduction}
\label{sec:introduction}
Universal machine-learning interatomic potentials (u-MLIPs) aim to achieve near-first-principles accuracy across broad chemical spaces without system-specific fitting. 
Leaderboard benchmarks such as MLIP Arena~\citep{chiang2026mlip} and Matbench Discovery~\citep{riebesell2025framework}, together with domain-specific benchmarks~\citep{krass2026mofsimbench,WELLENDORFF201536,loew2025universal}, are now primary evaluation tools, but evaluate models on predefined tasks covering only a limited region of configurational space.
Consequently, a model may score well yet behave unphysically elsewhere. Expanding benchmark coverage remains costly because each additional configuration requires a reference calculation, typically using density functional theory (DFT).

The challenge is therefore not to \emph{test more} but to decide \emph{where to test}.
To address this challenge, we formulate this problem as \emph{active failure mode discovery}: generating falsifiable failure hypotheses, screening them with inexpensive signals, and reserving costly DFT and expert verification for the most promising candidates.
Because this workflow demands open-ended reasoning over chemistry and simulation protocols, we cast it as a task for large language model (LLM) agents.
Related approaches have been explored for evaluating LLMs themselves~\citep{kossen2021activetesting,huang2026proeval,lin2025fact,tu2026benchguard,wang2026automated,wang2026androids,kossen2022active,berrada2026scaling,huang2026probellm}, but they do not directly transfer to MLIPs, whose evaluation relies on physical protocols and therefore presents a distinct setting. 
Proof-Carrying Materials (PCM) recently introduced adversarial auditing over predefined compositional descriptors, using an LLM as one of several proposal strategies for generating numerical feature vectors for a predefined evaluation oracle~\citep{basu2026proof}. To our knowledge, physics-informed active failure mode discovery for MLIPs remains largely unexplored. An extended related work discussion is provided in Appendix~\ref{app:related_work}.

We propose \textbf{MLIP Detective}, an agentic framework for active failure mode discovery in u-MLIPs.
Our framework uses existing benchmarks~\citep{riebesell2025framework, chiang2026mlip,krass2026mofsimbench,WELLENDORFF201536,loew2025universal} as a starting point for active search beyond their predefined coverage.
MLIP Detective does more than investigate where a model fails: it flags anomalous model behavior, proposes its physical cause, quantifies its downstream impact, and delivers a self-contained verification package (a report, reproduction code, and DFT-ready input structures) so that a human expert can confirm the finding independently. This human-agent collaboration enables automated exploration while preserving human final judgment and supporting trustworthy, auditable conclusions.

\section{Methodology}
\label{sec:mlip_detective}
\begin{figure}[t]
    \centering
    \includegraphics[width=\linewidth]{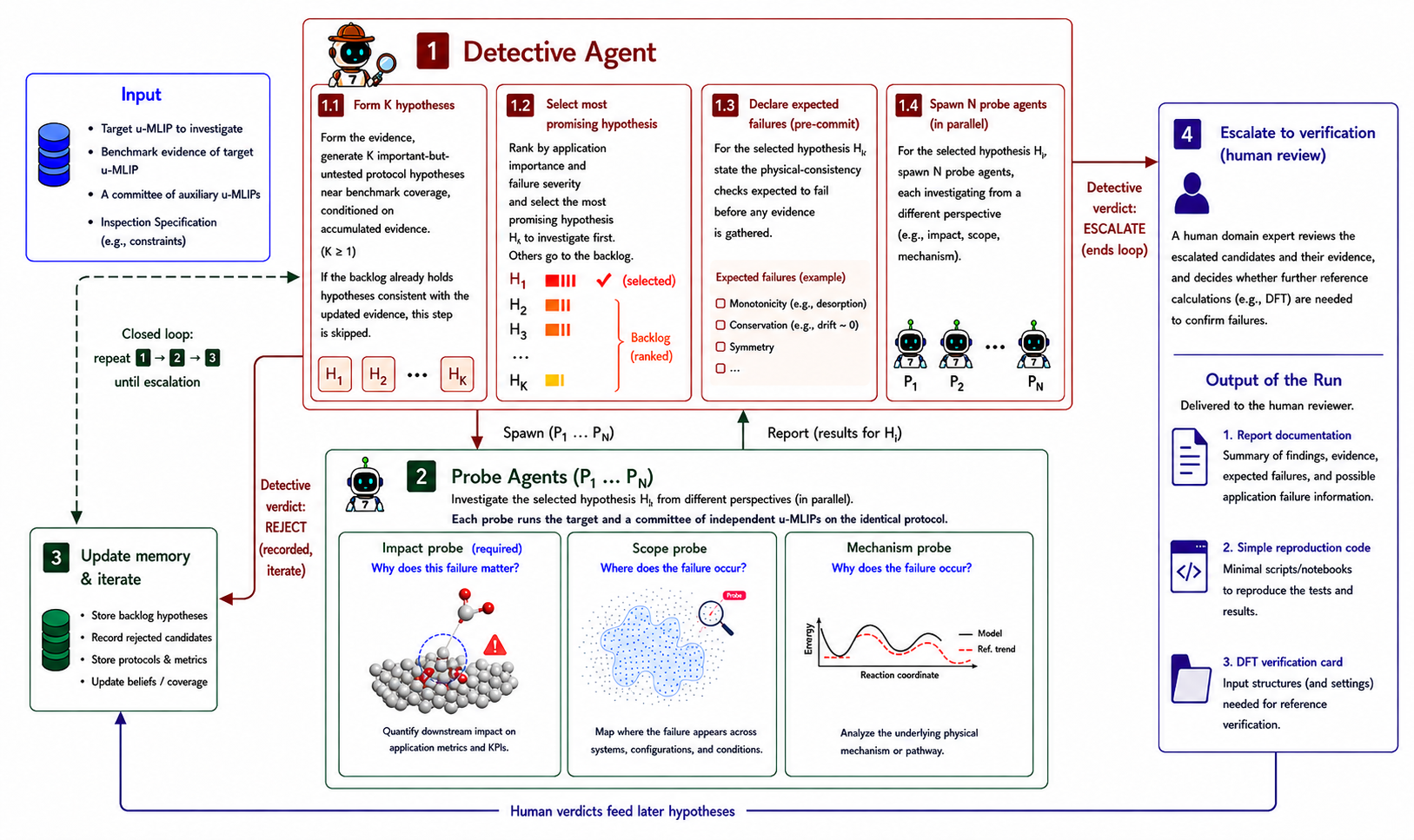}
    \caption{Overview of the MLIP Detective framework.}
    \label{fig:architecture}
    \vspace{-2mm}
\end{figure}

\textbf{Problem Setup.}
A target u-MLIP is a predictive model trained on reference data (e.g., DFT) that takes an atomic configuration as input and outputs quantities such as energy, forces, and stress. 
A \emph{test protocol} consists of related configurations together with an observable (e.g., a single-point energy). 
Failure mode discovery seeks protocols that expose candidate failures, such as spurious energy barriers.
A verifier (e.g., a human expert) determines whether each candidate constitutes a confirmed failure.
Confirmed failures are then grouped into \emph{failure modes}, which represent recurring, physically coherent patterns rather than isolated error points.

\textbf{MLIP Detective.}
The design of MLIP Detective for active failure mode discovery comprises three core components: (i) \textbf{surrogate signals} that estimate failure likelihood without requiring immediate costly verification, using cross-model disagreement and physical constraints; (ii) \textbf{LLM-driven failure hypothesis generation} that proposes physics-informed test protocols beyond existing benchmarks; and (iii) \textbf{an acquisition strategy} that prioritizes the most suspicious candidates for expert verification.

Figure~\ref{fig:architecture} provides an overview of MLIP Detective.
The framework has access to (i) the target u-MLIP, (ii) its results on existing benchmarks, (iii) a committee of separately trained auxiliary u-MLIPs, and (iv) a human-authored \emph{inspection specification}  that encodes physical constraints and diagnostic expectations and steers agents toward scientifically important or under-tested regions of configurational space.
The framework outputs a single candidate failure mode together with its test protocol, supporting evidence, reproduction code, and structures prepared for independent verification.
Two agent roles drive the investigation: a single \emph{Detective Agent} that generates candidate failure hypotheses and decides whether to escalate to human verification, and multiple \emph{Probe Agents} that execute computational probes. Agent instructions are given in Appendix~\ref{app:framework-details}.

In the MLIP Detective's search loop, (1) the Detective Agent generates and ranks candidate failure hypotheses based on application impact and severity, declares expected failures of the highest-priority hypothesis, and spawns parallel Probe Agents. (2) Probe Agents investigate the hypothesis from multiple perspectives, e.g., why the failure matters, where it occurs, and why it occurs. They run their protocols independently on the target and committee models and return analyzed reports to the Detective Agent.
(3) The Detective Agent evaluates each report using two label-free surrogate signals: cross-model disagreement and violations of the constraints documented in the specification. All reports, including rejected failure hypotheses, are stored in an \emph{evidence memory} that informs later iterations. If any hypothesis carries sufficiently strong surrogate evidence, the Detective Agent \emph{escalates} it to verification and the search ends; otherwise it updates the backlog and returns to (1).

\begin{figure*}[t]
\centering
\includegraphics[width=\textwidth]{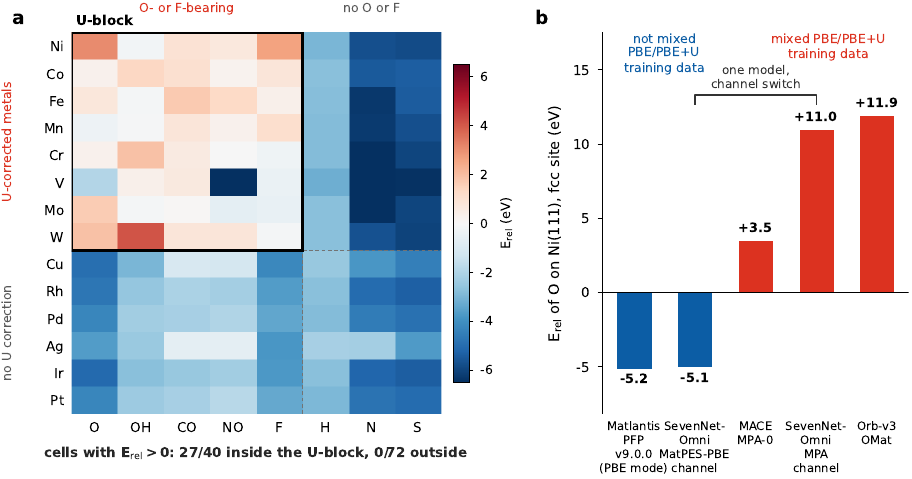}
\caption{
\textbf{MACE-MPA-0 predicts anomalous energies only in the $U$-block.}
\textbf{a}, Relative energy \(E_{\mathrm{rel}}\) of MACE-MPA-0 across 14 metal surfaces and 8 adsorbates, referenced to the clean slab and isolated adsorbate.
\(E_{\mathrm{rel}}>0\) indicates that the relaxed adsorbate--surface system lies above its separated-fragment reference; MLIP Detective used \(E_{\mathrm{rel}}>0\) as a diagnostic flag for candidate failures.
All 27 anomalous cases lie within the outlined $U$-block (27 of 40 cells).
\textbf{b}, \(E_{\mathrm{rel}}\) for O on Ni(111) at identical geometry across five model/channel configurations.
The prediction varies substantially across model channels, supporting an association between the anomaly in the $U$-block and mixed PBE/PBE+$U$ training data.}
\label{fig:useam}
\vspace{-4mm}
\end{figure*}

\begin{table*}[t]
\centering
\caption{Action log for one MLIP Detective search iteration, organized by the steps in Figure~\ref{fig:architecture}.}
\label{tab:detective_action_log}
\footnotesize
\setlength{\tabcolsep}{3.5pt}
\renewcommand{\arraystretch}{0.90}
\setlength{\aboverulesep}{0.25ex}
\setlength{\belowrulesep}{0.25ex}

\begin{tabularx}{\textwidth}{@{}
>{\centering\arraybackslash}p{0.7cm}
>{\raggedright\arraybackslash}p{3.0cm}
X
@{}}
\toprule
\textbf{Step} & \textbf{Action} & \textbf{Description} \\
\midrule

1.1
& \textbf{Form $K$ hypotheses}
& The Detective Agent computed a benchmark MAE of $1.21$~eV against
DFT-PBE. The errors were bimodal: 5 of 36 systems showed large errors,
all involving Ni or Co with oxygen-bearing adsorbates. The relaxed
O/Ni(111) adsorbate--slab system retained a 2.05~\AA{} bond but had
$E_{\mathrm{rel}}=+3.09$~eV. The Detective Agent formed three
hypotheses concerning an anomaly in the $U$-block, CO overbinding on
noble metals, and physisorption on Pt. \\
\midrule

1.2
& \textbf{Select most promising hypothesis}
& The Detective Agent ranked the three candidate failure hypotheses by
application impact and severity, selected the hypothesis concerning
the $U$-block, and recorded the other two for later iterations. \\
\midrule

1.3
& \textbf{Declare expected failures (pre-commit)}
& The Detective Agent refined and pre-committed the selected failure hypothesis:
a mixed PBE/PBE+$U$ reference inconsistency affecting combinations in
the $U$-block. The hypothesis predicted
the same anomaly for six unseen metals and for F but not S, consistent
force--energy behavior, and no anomaly outside the $U$-block. \\
\midrule

1.4
& \textbf{Spawn $N$ Probe Agents}
& The Detective Agent launched three Probe Agents to test the scope,
mechanism, and application impact of the selected hypothesis. \\
\midrule

2
& \textbf{Probe Agents ($P_1$--$P_3$, parallel)}
& \textbf{Scope Probe ($P_1$):} evaluated the predicted scope of the
hypothesis using a $14\times8$ metal--adsorbate grid across five
model/channel configurations. MACE-MPA-0 gave $E_{\mathrm{rel}}>0$ for
27 of 40 combinations within the $U$-block and 0 of 72 outside it.
The anomaly appeared for F but not S, and for nonmagnetic Mo and W but
not Cu, making oxygen-specific, magnetic, and 3d-only explanations less likely.
\textbf{Mechanism Probe ($P_2$):} found a metastable minimum 3.5~eV above the separated asymptote and found no evidence that force inconsistency or dispersion explained the anomaly.
\textbf{Impact Probe ($P_3$):} found that 4 of the 5 O-bearing steps in the 10-step methanation chain were shifted by 2.4--5.1~eV and that the predicted Ni(111) oxidation and CO-coverage thermodynamics were qualitatively incorrect at operating conditions, while a short low-coverage MD run remained indistinguishable from the corresponding committee runs. \\
\midrule

3
& \textbf{Update memory \& iterate}
& After reviewing the three Probe Agent reports, which independently identified the selected SevenNet-Omni channel as a key variable, the Detective Agent re-derived all load-bearing quantities from the probes' raw outputs, updated the evidence memory, and reframed the finding as a corpus-associated pattern. Each Probe Agent had autonomously added a \textbf{within-model channel-switch experiment}; in the Scope Probe's report, switching SevenNet-Omni from its MatPES-PBE channel to its MPA channel changed the mean $E_{\mathrm{rel}}$ by $+4.8$~eV inside the $U$-block but only $-0.02$~eV outside it. \\
\midrule

4
& \textbf{Escalate to verification (human review)}
& The Detective Agent prepared a minimal reproduction, 11 plane-wave
DFT input configurations, and a decision table mapping possible DFT
outcomes to verdicts. The search loop ended at escalation. \\

\bottomrule
\end{tabularx}
\end{table*}

\section{Experimental Results}
\label{sec:results}

We present two complementary case studies on MACE-MPA-0~\citep{batatia2023foundation}. Section~\ref{sec:u_seam} shows how the current MLIP Detective, while inspecting a single model, rediscovered adsorption-energy anomalies associated with selective training data and localized its manifestation across chemical space. Section~\ref{sec:co_cu} shows how an earlier prototype identified a blind spot in an adsorption-energy benchmark, discovered a model defect along an untested desorption pathway, and escalated it for verification. Plane-wave DFT single-point calculations on the same path geometries showed no comparable barrier along the desorption path. An additional finding for SevenNet-Omni is provided in Appendix~\ref{app:additional-results}.
Claude Opus~5 was used for the case study in Section~\ref{sec:u_seam}, and Claude Opus~4.8 was used for Section~\ref{sec:co_cu} as the underlying LLM, respectively.

\subsection{\texorpdfstring{Rediscovery of Adsorption-Energy Anomalies Associated with Selective PBE+$U$}{Rediscovery of Adsorption-Energy Anomalies Associated with Selective PBE+U}}
\label{sec:u_seam}
\textbf{The finding.}
In this run, we inspected MACE-MPA-0 (trained on MPtrj~\citep{deng_2023_chgnet}+sAlex~\citep{Barros-Luque2026}) against a committee of three u-MLIPs: the latest version, v9.0.0, of Matlantis-PFP~\citep{matlantis,10.1038/s41467-022-30687-9}, SevenNet-Omni~\citep{kim2026optimizing}, and Orb-v3~\citep{rhodes2025orb}. The input benchmark evidence is a 36-system subset of the metal-surface adsorption-energy benchmark compiled by ~\citet{WELLENDORFF201536}.\footnote{We excluded reactions 24, 25, and 39 of Table 3 in the original study: reactions 24 and 25 lack a single clean reference reaction, and reaction 39 is a differential reaction on a pre-covered surface.}
We define the $U$-block as the metal--adsorbate combinations that trigger the Materials Project's selective Hubbard $U$ correction~\citep{mp_hubbard_u}: one of the designated transition metals paired with an O- or F-bearing adsorbate.
MLIP Detective independently surfaced a recently reported training-data inconsistency~\citep{warford2026betteruimpactselective,kim2026optimizing}: mixing PBE with selectively applied PBE+$U$ reference energies creates incompatible potential energy surfaces, producing anomalous adsorption energetics for combinations of elements within the $U$-block.
This case demonstrates the framework's ability to turn benchmark evidence into a localized, falsifiable failure hypothesis and a specific training-data attribution.\footnote{The run inputs and logs contain no indication that the issue was provided or retrieved, although prior exposure through LLM pretraining cannot be excluded.}
We define the energy of the combined adsorbate--slab system relative to the separated fragments as
$E_{\mathrm{rel}}\equiv E_{\mathrm{combined}}
-E_{\mathrm{clean\,slab}}-E_{\mathrm{isolated\,adsorbate}}$;
$E_{\mathrm{rel}}>0$ means that the relaxed adsorbate--slab system lies above the corresponding separated fragments.
Across a 14-metal $\times$ 8-adsorbate grid, 27 of the 40 combinations within the $U$-block yield $E_{\mathrm{rel}}>0$, whereas none of the 72 combinations outside the $U$-block do (Fig.~\ref{fig:useam}a).

The Detective Agent converted this probe evidence into an attribution analysis and identified the likely cause: training on mixed PBE/PBE+$U$ data, which introduces an inconsistent energy reference between PBE-labeled and PBE+$U$-labeled systems.
Cross-model comparisons and the within-model SevenNet-Omni channel switch in Fig.~\ref{fig:useam}b support the hypothesis that mixed PBE/PBE+$U$ reference energies are the likely origin of the anomaly.
Because closely related prior work~\citep{kim2026optimizing,warford2026betteruimpactselective} has already documented the same issue, the human verifier judged that additional DFT calculations were unnecessary for this case.

\textbf{Search trajectory.}
Table~\ref{tab:detective_action_log} summarizes the logs from hypothesis generation to escalation. 

\subsection{DFT-Verified Anomaly beyond Adsorption-Energy Benchmark Coverage}
\label{sec:co_cu}
\textbf{The finding.}
In this run, we evaluated \mbox{MACE-MPA-0} against a committee of Matlantis-PFP~v8.0.0, Orb-v3, and SevenNet~\citep{park_scalable_2024} using the same 36-system subset of the metal-surface adsorption-energy benchmark~\citep{WELLENDORFF201536} used in Section~\ref{sec:u_seam}. The benchmark provides both experimental adsorption energies and DFT-PBE values. Section~\ref{sec:u_seam} reports the MAE against DFT-PBE, whereas the 0.04~eV comparison reported below uses the experimental reference at the on-top site.

What the Detective Agent exploited is a blind spot of that benchmark:
it scores the adsorption energy only at a single relaxed adsorption
geometry, so
it constrains the \emph{depth} of the adsorption well and says nothing
about the \emph{shape} of the potential energy surface along the
desorption coordinate.
CO/Cu(111) shows how far apart the two can be.
At the on-top site, where the benchmark compares it with experiment,
MACE-MPA-0 reproduces the experimental adsorption energy of CO to
within $0.04$~eV, but the benchmark never evaluates the desorption path
from that site.
A Probe Agent investigating the shape of the adsorption PES performed a
rigid scan along this path, translating CO along the surface normal
away from the relaxed on-top geometry:
MACE-MPA-0 rises $0.22$~eV \emph{above} its own desorption asymptote
at $\Delta z \approx 1.25$~\AA{} before dropping into a
secondary well of $-0.054$~eV at $\Delta z \approx 2.75$~\AA{}.
None of the three committee models ever exceeds its asymptote.
Table~\ref{tab:co_cu_paths} lists this scan and the calculations of
the human verification below, with the site each starts from.

\begin{table}[b]
\centering
\caption{Calculations on the CO/Cu(111) desorption path and the
adsorption site each starts from. Values are the maximum energy (eV)
relative to the desorbed state, taken as the last image of the band or
the last point of the scan, where CO is lifted by 5.5~\AA{} (11-image CI-NEB),
12~\AA{} (25-image CI-NEB), or 5.9~\AA{} (rigid scan) in the same cell. In the 11-image CI-NEB,
all six models start from one common on-top geometry; in the 25-image
CI-NEB, each model starts from its own lowest-energy site.}
\label{tab:co_cu_paths}
\small
\begin{tabular}{@{}llrlr@{}}
\toprule
Calculation & Start site & MACE-MPA-0 & Other models & DFT \\
\midrule
Rigid scan, 4 models & on-top & $+0.22$ & committee: none above & --- \\
\quad (Probe Agent) & & & & \\
CI-NEB, 11 images, 6 models & on-top & $+0.188$ & UMA-S-1.2: $+0.063$ & --- \\
 & & & MatterSim: $+0.053$ & \\
 & & & committee: none above & \\
CI-NEB, 25 images, 2 models & fcc hollow & $+0.208$ & --- & $+0.022$ \\
 & on-top & --- & Matlantis-PFP: none above & none above \\
\bottomrule
\end{tabular}
\end{table}

\textbf{Human verification.}
To rule out a constrained-scan artifact, the human verification step
began with a preliminary check on minimum-energy paths, using
climbing-image nudged elastic band (CI-NEB)
calculations~\citep{henkelman2000climbing} in which each model relaxed
its own endpoints from one common geometry at the on-top site of the
scan, that of Matlantis-PFP~v8.0.0 in the benchmark.
Each band had 11 images, with the desorbed endpoint lifted by
5.5~\AA{}, and was optimized with FIRE~\citep{bitzek2006structural}
until the maximum NEB force fell below 0.10~eV\,\AA$^{-1}$;
the other settings followed the 25-image CI-NEB described in the
computational verification details below.
MACE-MPA-0 rises $+0.188$~eV above the desorbed state on CO/Cu(111),
whereas none of the three committee models ever exceeds it.
Two models outside the committee, UMA-S-1.2 (OC20 task)~\citep{wood2026family}
and MatterSim~\citep{yang2024mattersim}, also rise above it, by
smaller amounts, $+0.063$ and $+0.053$~eV, respectively.
Over the five other systems of the family, computed in the same way,
MACE-MPA-0 rises above the desorbed state on NO/Pd(100) ($+0.117$~eV),
NO/Pd(111) ($+0.095$~eV), and NO/Pt(111) ($+0.031$~eV, below Orb-v3's
$+0.048$~eV), and not on CO/Ru(001) or CO/Ir(111).
Across the six systems, no other model rises more than $+0.063$~eV
above the desorbed state.

\begin{figure}[t]
\centering
\includegraphics[width=\linewidth]{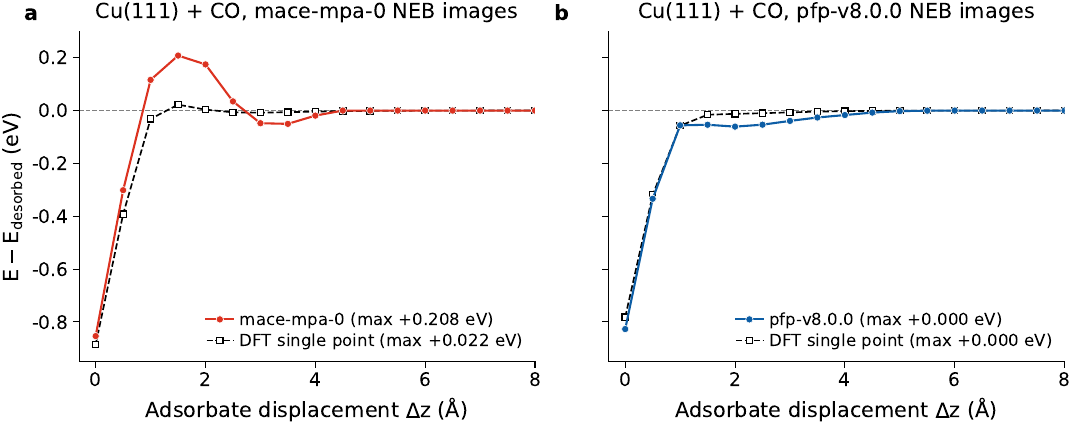}
\caption{\textbf{DFT verification of the CO/Cu(111) desorption
barrier.} Energy relative to the desorbed state along the
CI-NEB path, for \textbf{a},
MACE-MPA-0 and \textbf{b}, Matlantis-PFP~v8.0.0. Filled circles are
the model's own band; open squares are plane-wave DFT (PBE) single
points on the same 25 image geometries. Each NEB band was initialized with 25 images, including both endpoints, at 0.5~\AA{} intervals over a total desorption displacement of 12~\AA{}. Only the range $\Delta z$ = 0–8~\AA{} is shown.
MACE-MPA-0 places a
$+0.208$~eV maximum above the desorbed state (dashed line; CO lifted by 12~\AA{} in the same cell) where DFT
gives $+0.022$~eV; on the Matlantis-PFP path neither curve exceeds its desorbed-state reference.}
\label{fig:co_cu_dft}
\end{figure}

Because MACE-MPA-0 rises most on CO/Cu(111), its behavior on this system was then
compared with DFT on a separate CI-NEB with tighter settings (25
images, maximum NEB force 0.05~eV\,\AA$^{-1}$), run for MACE-MPA-0
and, as a control, for Matlantis-PFP~v8.0.0, each starting from its
own lowest-energy adsorption site (fcc hollow for MACE-MPA-0, on-top
for Matlantis-PFP).
Plane-wave DFT-PBE single-point calculations on every image of the
two converged paths show that MACE-MPA-0 substantially overestimates
the rise (Fig.~\ref{fig:co_cu_dft}): along its path, the model rises
$+0.208$~eV above the desorbed state, whereas the DFT reference rises
only $+0.022$~eV, an order of magnitude smaller.
On the Matlantis-PFP path, neither the model nor the reference ever
exceeds the desorbed state.

\textbf{Computational verification details.}
The computational setup used to obtain the energy profiles in Fig.~\ref{fig:co_cu_dft} is described below.
For each of MACE-MPA-0 and Matlantis-PFP v8.0.0, the CO desorption path was optimized using the climbing-image nudged elastic band (CI-NEB) method implemented in the Atomic Simulation Environment (ASE)~\citep{ase-paper}, with energies and forces evaluated by the respective MLIP. The simulation cell consisted of a four-layer $2\times2$ Cu(111) slab and one CO molecule, with approximately 42~\AA{} of vacuum normal to the surface. All Cu atoms were held fixed during both endpoint relaxation and NEB optimization.

Each model started from its own lowest-energy adsorption geometry.
The endpoints were relaxed separately with each MLIP using FIRE~\citep{bitzek2006structural}. Each NEB band contained 25 images, including both endpoints, and was initialized using image dependent pair potential (IDPP) interpolation~\citep{smidstrup2014improved}. The bands were optimized with FIRE, using a uniform spring constant of 0.1~eV\,\AA$^{-2}$ and a maximum NEB-force convergence threshold of 0.05~eV\,\AA$^{-1}$.

The converged NEB images were evaluated directly, without further interpolation, by single-point DFT calculations using VASP~6.5.1~\citep{kresse1993ab,kresse1996efficient} with the Perdew–Burke–Ernzerhof (PBE) functional~\citep{perdew1996generalized} and projector augmented-wave (PAW) potentials~\citep{kresse1999ultrasoft} from the PBE\_64 library: \texttt{PAW\_PBE Cu\_pv 06Sep2000}, \texttt{PAW\_PBE C 08Apr2002}, and \texttt{PAW\_PBE O 08Apr2002}. 
The gas-phase CO reference was obtained by relaxing an isolated
molecule in a cubic cell with a side length of 15~\AA{}.
Spin-polarized collinear DFT calculations employed a plane-wave cutoff
of 680~eV, a $7\times7\times1$ $k$-point mesh for the slab cell and
$2\times2\times2$ for the molecular box, Gaussian smearing with
$\sigma = 0.05$~eV, and an SCF energy convergence threshold of
$10^{-6}$~eV.

\textbf{Relation to the framework of this paper.}
This finding was produced by the first prototype of MLIP Detective,
which differs from the architecture defined in
Section~\ref{sec:mlip_detective} in four respects.
(i)~There was no \emph{evidence memory} and no feedback path into it.
(ii)~A hypothesis was tested by a \emph{single} probe rather than by
several probes instantiating different variations of it.
The complementary perspectives that the current framework requires in
parallel, in particular the impact probe that bounds where the defect
does and does not matter, were therefore absent.
(iii)~Several hypotheses were pursued \emph{concurrently} rather than
one at a time.
The run spawned eight probes across independent directions (elastic
and phonon stability, short-range repulsion, clean-metal and
oxide/semiconductor surface energies, adsorption PES shape), of which
one produced the finding above and one a low-severity finding.
(iv)~The format of the output handed to the human verifier was free-format, and reviewing them took correspondingly more human effort.
Together, these lessons shaped the more systematic and auditable workflow described in Section~\ref{sec:mlip_detective}.

\section{Limitations and future work}
\label{}
Our study demonstrates that an agentic framework can be useful for physics-informed active failure mode discovery. 
However, the search remains bound by the inspection specification, and escalated findings require costly human verification (see Appendix~\ref{app:limitations} for more details on limitations). 
Future work includes automating the quality control of escalated findings so that experts focus only on the most impactful candidates, and extending the search to broader classes of simulation protocols and target models.

\section{Acknowledgments}
\label{sec:acknowledgments}
We thank our colleagues at Preferred Networks, Inc.: So Takamoto and Kohei Shinohara for discussions that helped inspire the initial idea for this work, and Chikashi Shinagawa and Katsuhiko Nishimra for their advice on and assistance with the reference DFT calculations.

\bibliographystyle{plainnat}
\bibliography{references}


\appendix

\section{Related work}
\label{app:related_work}

\paragraph{Benchmarks for u-MLIPs.}
Community evaluation of u-MLIPs is dominated by benchmark suites, including Matbench Discovery~\citep{riebesell2025framework}, MLIP Arena~\citep{chiang2026mlip}, and domain-specific benchmarks~\citep{WELLENDORFF201536,krass2026mofsimbench,loew2025universal}.
These suites are the de facto standard for model comparison and are also the starting point of our framework: MLIP Detective consumes their results as evidence and extends inspection into the configurations they leave untested. Our work is thus complementary to benchmark development rather than a replacement.

\paragraph{Active testing and efficient evaluation.}
Active testing formalizes model evaluation under a costly labeling oracle: the model is fixed, and test inputs are selected to make the most of each label~\citep{nguyen2018active, kossen2021activetesting, kossen2022active}. Surrogate estimates of expected loss guide importance-sampling acquisition, while the LURE estimator corrects for the resulting selection bias; follow-up work applies related ideas to LLM evaluation~\citep{berrada2026scaling}.
These methods operate on a static pool and target unbiased estimation of an aggregate metric. Our problem shares the costly-oracle premise and the fixed-model setting, but differs from classical active testing in targeting failure mode discovery rather than unbiased estimation and in generating candidates beyond any fixed test pool.

\paragraph{Proactive evaluation of generative models.}
ProEval~\citep{huang2026proeval}, the closest analogue to our setting, frames performance estimation and failure discovery as dual Bayesian objectives. It uses transfer learning to construct an informed Gaussian process (GP) prior from historical evaluation results or semantic embeddings, then applies Bayesian quadrature to performance estimation and superlevel set sampling (SS) to failure discovery. SS is extended through generative synthesis (SS-Gen), which uses identified hard examples as in-context anchors, and topic-aware exploration (TSS), which promotes diverse failure patterns. Building on Bayesian level set estimation~\citep{gotovos2013lse}, ProEval contributes transfer learning for GP priors and active synthesis of new test cases. MLIP Detective instantiates analogous components for MLIPs: cross-model disagreement and physical-constraint violations replace the GP surrogate; agents construct physics-informed test protocols from benchmark evidence; and the oracle is an expert verdict supported by costly DFT.

\paragraph{Adaptive probing of LLMs.}
Fact-Audit~\citep{lin2025fact} uses importance sampling and an adaptive scenario taxonomy to focus fact-checking probes on model weaknesses. ProbeLLM~\citep{huang2026probellm} uses hierarchical MCTS to balance MACRO coverage and MICRO refinement, then derives failure modes using failure-aware embeddings and boundary-aware induction. MLIP Detective transfers this adaptive probing paradigm to MLIPs through physics-informed test protocols and expert verification supported by DFT.

\paragraph{Red teaming.}
Automated red teaming seeks inputs that trigger undesired behavior, via attacker fine-tuning~\citep{perez2022red}, iterative refinement~\citep{chao2025jailbreaking, mehrotra2024tree}, or quality-diversity search~\citep{samvelyan2024rainbow}. These methods optimize for eliciting failures, typically without a per-query verification oracle of DFT-like cost, and diversity is enforced heuristically. Our setting differs in that many candidate findings require costly DFT calculations for confirmation.

\paragraph{Uncertainty quantification and active learning for MLIPs.}
Within the MLIP community, ensemble variance and related uncertainty estimates guide reference-data acquisition in active-learning loops~\citep{behler2016perspective,podryabinkin2017active,smith2018less}, while adversarial sampling targets high-uncertainty, thermally accessible configurations for retraining~\citep{schwalbe2021differentiable,cezar2025learning}. Like MLIP Detective, these methods use inexpensive surrogate signals to prioritize costly reference calculations. Their goal, however, is to update the model; ours is to evaluate a fixed model.

\paragraph{Adversarial auditing of MLIPs.}
Proof-Carrying Materials (PCM)~\citep{basu2026proof} searches for compositional blind spots in a predefined tabular feature space. Its LLM is one of six adversaries and proposes numerical feature vectors, which the oracle maps to existing materials for MLIP--DFT evaluation. MLIP Detective instead searches over physical test protocols comprising related configurations, observables, and falsifiable consistency criteria. This protocol-level search can expose relational failures, including unphysical dissociation limits, barriers, and force--energy inconsistencies, that are not naturally represented as isolated points in a fixed descriptor space. It also enables agents to generate new configurations, localize the conditions under which an anomaly occurs, and produce a physically interpretable hypothesis with a DFT-ready verification protocol. Cross-MLIP disagreement and physical-constraint violations are used to prioritize which candidate failure hypotheses receive new DFT calculations and expert review.

\paragraph{Agentic frameworks for atomistic simulation.}
Recent work applies LLM agents to MLIP workflows, e.g., MLIPilot~\citep{osaro2026mlipilot}, OptiMat Alloys~\citep{hu2026optimat}, and Lang2MLIP~\citep{li2026lang2mlip}, automating simulation setup or model construction. MLIP Detective differs in purpose: the agents are used not to run simulations for a user's scientific goal but to inspect the model itself, forming and testing hypotheses about where it fails.

\section{Framework and implementation details}
\label{app:framework-details}
This appendix documents the operational details of MLIP Detective: the
requirements under which the agents operate, the inspection
specification, and the implementation substrate.

The Detective Agent operates from a human-authored
natural-language instruction document, fixed before any run; each
Probe Agent operates from a brief that the Detective Agent composes at spawn
time.
This appendix excerpts the requirements that define each component of
the framework.
The boxed passages (Boxes~\ref{box:roles}--\ref{box:signals}) and
inline quotations are verbatim
from the operational instructions, with four classes of edits: role
and artifact names are mapped to the paper's terminology, references to
file paths and internal infrastructure are replaced by bracketed
placeholders, omissions are marked by ellipses in brackets, and markup
is converted from the source's plain-text formatting.
No physics content was added, removed, or reworded.

\paragraph{Roles and authority.}
The instructions open by fixing the division of authority between the
Detective Agent, the Probe Agents, and the human verifier
(Box~\ref{box:roles}). Two constraints of Section~\ref{sec:mlip_detective}
are made explicit here: the Detective Agent's verdict vocabulary
contains no \textsc{confirm}, and no DFT call is permitted inside the
search loop.

\begin{promptbox}[box:roles]{Division of authority (The Detective Agent instructions)}
The division of authority is strict:
\begin{itemize}
  \item The detective forms hypotheses, declares expected failures,
  spawns probes, and decides which candidates to \textsc{reject} or
  \textsc{escalate}.
  \item It can never \textsc{confirm} a failure. Confirmation happens
  outside the loop: the human operator [\dots] renders the verdict as a
  domain expert.
  \item Inside the loop everything is label-free: no DFT is called. DFT
  belongs to the post-loop verification stage and is budget-bounded
  (one call per configuration in an escalated protocol).
\end{itemize}
\end{promptbox}

\paragraph{Detective Agent: hypothesis discipline and pre-registration.}
The Detective Agent is required to pursue one hypothesis at a time:
``Hypotheses are pursued one at a time: settle one, explore it, and
move to the next only after the current one is settled (escalated,
rejected, or exhausted). Each new hypothesis is conditioned on the full
evidence memory, including the outcome of the one before it.''
Candidate directions are ranked by the two criteria of
Section~\ref{sec:mlip_detective}:
\emph{importance} (``how much the region matters in practical
applications of [the target]'') and \emph{severity} (``how badly the
model would be wrong if the hypothesis holds (large spurious barriers,
wrong phase ordering, unphysical forces---not marginal numerical
noise)'').

Before any Probe Agent is spawned, the Detective Agent must pre-register the
expected failure (Box~\ref{box:declaration}). This declaration serves
three functions in the current framework. First, it constrains the
Detective Agent's own later review step: because the Detective Agent both generates
hypotheses and judges the Probe Agents' evidence, the pre-registered checks
prevent apparent anomalies in probe output from being rationalized
post hoc into ``the expected failure''. Second, it commits the
hypothesis to outcomes on configurations not yet evaluated, which is
what makes a discovery claim verifiable. Third, it is the source of
the decision table in the escalation report
(see below), so the human verdict is an
application of pre-committed rules.

\begin{promptbox}[box:declaration]{Pre-registration requirement (Detective Agent instructions)}
Before spawning anything, append to [the evidence memory]:
\begin{itemize}
  \item run id (timestamp) and target model,
  \item the hypothesis [\dots] and the committee list,
  \item the \emph{declaration}: for each probe variation you are about
  to spawn (structure subset / observable), the observable and the
  concrete physical-consistency checks you expect the target model to
  fail (e.g.\ no spurious extremum along the path, correct force
  directions, correct energetic ordering), written verbatim
  \emph{before} any evidence is gathered.
\end{itemize}
The declaration does not define failure---the human verdict [\dots]
does. It documents what was predicted, gives the expert
a rubric to judge against [\dots].

The declared variations \emph{must} include at least one \emph{impact
probe}: a variation that measures the hypothesized defect's consequence
in an application-like setting (condensed phase, realistic workflow,
finite temperature) rather than the idealized geometry that exposes it.
Its result feeds the ``application failure'' section of the escalation
report [\dots]---and a probe showing the defect is screened away in
practice is grounds to reject or de-prioritize, not escalate.
\end{promptbox}

\paragraph{Detective Agent: evidence memory and information hygiene.}
The evidence memory is an append-only file recording declarations,
probe reports, rejections with their reasons, and post-loop human
verdicts. The instructions restrict what a new run may read from past
runs: full reports of settled runs are archived out of the agents'
reach, and the distilled evidence memory is ``the only carry-over from
past runs---reading old reports would bias new hypothesis
formation''.

\paragraph{Probe Agent briefs.}
Each Probe Agent receives a brief composed by the Detective Agent, stating the
target and committee models, the hypothesis, this Probe Agent's assigned
variation of it, and the declared checks. Three further requirements
on the brief matter for the results of this paper. First, the
committee obligation: the Probe Agent ``\emph{must} run the identical
protocol on the target \emph{and} every committee model, and report
both surrogate signals'', and ``if every model, target and committee
alike, shows the same anomaly, the protocol rather than the model is
suspect''. Second, methodological freedom: the brief must not
prescribe a fixed list of check functions; the probe ``is expected to
design its own diagnostics around the hypothesis'', under the design
rule ``be specific in the hypothesis; be open in the methodology''.
Third, the report format: a self-contained report whose protocol (the
atomic configurations as files, plus the observable computed from
them) is ``re-runnable as-is'', with per-model observable values, the
cross-model disagreement, the physical-constraint checks run, and
severity as a continuous quantity (e.g.\ the height of a spurious
barrier).

\paragraph{Evaluation and escalation.}
On receiving a probe report, the Detective Agent reviews it against the
inspection specification, the target model's documentation, and the
pre-registered declaration, and renders \textsc{reject} (appending the
reason to the evidence memory) or \textsc{escalate}. The instructions
fix the semantics of the two surrogate signals asymmetrically
(Box~\ref{box:signals}): an exact-constraint violation is decisive on
its own, whereas disagreement only allocates suspicion, and committee
unanimity is neither expected nor required.

\begin{promptbox}[box:signals]{Semantics of the surrogate signals (Detective Agent instructions)}
The two signals establish different things. A violation of a physical
constraint that applies exactly to the observable is decisive on its
own---the inconsistency is established without any reference.
Cross-model disagreement is a suspicion signal: at least one model must
be wrong, but disagreement alone cannot say which. Committee unanimity
is therefore neither expected nor required. Use physical constraints
and physical intuition---not mere outlier-ness in value---to decide who
the prime suspect is:
\begin{itemize}
  \item If the \emph{target} model departs furthest from the physics,
  the region is worth exploring even when committee members also behave
  oddly there. [\dots]
  \item If a \emph{committee} member shows the more pronounced anomaly,
  it is not the current run's finding: record it in [the evidence
  memory] as a future study target [\dots], and judge the current
  target against the remaining members.
\end{itemize}
\end{promptbox}

An escalation must be delivered as a fixed-format escalation report:
the document the human verifier reads. Its required sections
are:
(1)~\emph{Finding}, one to two sentences plus the benchmark evidence
the hypothesis started from; (2)~\emph{Type}, exact-constraint
violation or cross-model disagreement, which determines whether
verification supplies reference values or decides the verdict;
(3)~\emph{Application failure}, grounded in the impact probe's
measurement and required to also state the domain boundary ``where the
defect does \emph{not} matter''; (4)~a single figure ``from which the
anomaly is visible at a glance''; (5)~a \emph{minimal reproduction}
script that ``runs in minutes and prints the headline numbers'';
(6)~a \emph{DFT verification card}: the configurations to compute with
their cost, ready-to-run inputs, and ``a decision table mapping DFT
outcomes to verdicts (`reference shows X $\rightarrow$ confirmed; shows
Y $\rightarrow$ refuted')'', derived from the declaration of
Box~\ref{box:declaration}; and (7)~\emph{Confounds closed}, at most
five bullets listing the artifact explanations tested and excluded.

\paragraph{The inspection specification.}
The inspection specification  of Section~\ref{sec:mlip_detective} is
a standalone document of physical constraints and diagnostic
expectations for MLIP validation. It
defines eight constraints: energy–force consistency ($F=-\nabla E$),
translational invariance / zero net force for an isolated system, a repulsive
wall at short range, elastic (Born) stability of known stable crystals,
symmetry preservation, stress--strain consistency, the dissociation
limit (size consistency at large separation), and PES smoothness. Each
comes with a statement of the constraint, its physical rationale, a
label-free check procedure, and a numerical threshold. Some of these
properties are guaranteed by construction in a given MLIP architecture
(e.g.\ force conservation via automatic differentiation); the
specification guides agents to skip such checks for the target model
and to concentrate on the properties that are not guaranteed.

Neither the inspection specification nor the framework instructions
refer to any specific material system, dataset, or known failure mode
of the target models. The remaining human-authored inputs to the run
of Section~\ref{sec:u_seam}---the seed benchmark with the target
model's results on it, the identity of the target and committee
models, and the documentation of the models and of the validation
suite that the instructions direct agents to consult---likewise
contain no reference to the pathology rediscovered there.

\paragraph{Implementation.}
Both roles are instances of a general-purpose LLM coding agent
(Claude Code), operating in a sandboxed workspace with file-system,
shell, and Python tools with access to the target and committee
MLIPs.
The agents are driven by the Claude Opus model available
at the time of each run: Claude Opus~4.8 for the run of
Section~\ref{sec:co_cu}, and Claude Opus~5 for the runs of
Section~\ref{sec:u_seam} and Appendix~\ref{app:additional-results}. The
Detective Agent and each Probe Agent run as separate agent sessions; the Detective Agent
spawns Probe Agents in parallel within a hypothesis, and each Probe Agent returns
its report to the Detective Agent.

\section{Additional Experimental Result: Mid-range dimer attraction in SevenNet-Omni}
\label{app:additional-results}

\begin{figure}[t]
\centering
\includegraphics[width=\linewidth]{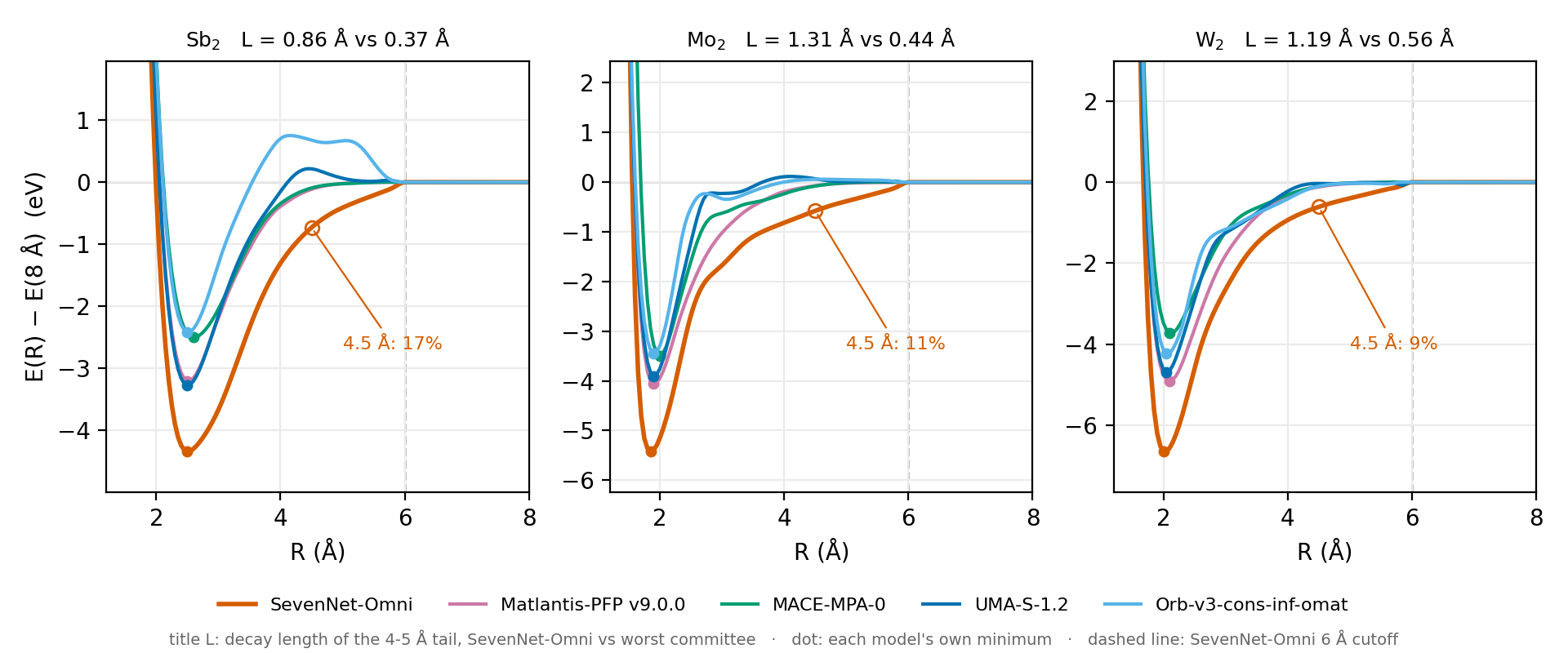}
\caption{\textbf{Anomalous mid-range attraction in SevenNet-Omni
dimers.} Binding curves $E(R) - E(8~\text{\AA})$ for Sb$_2$, Mo$_2$,
and W$_2$ computed with SevenNet-Omni (orange) and the four
committee models (Matlantis-PFP~v9.0.0, MACE-MPA-0, UMA-S-1.2, and
Orb-v3-cons-inf-omat).
Dots mark each model's own minimum; the dashed line marks
SevenNet-Omni's 6~\AA{} cutoff.
Annotations give the fraction of SevenNet-Omni's own well depth
still bound at 4.5~\AA{}, and each panel title reports the fitted
decay length of the 4--5~\AA{} tail for SevenNet-Omni versus the
worst committee model.}
\label{fig:omni_dimer_tail}
\end{figure}

This run demonstrates that the framework is not tied to a particular
target: here the roles are rearranged, with SevenNet-Omni~\citep{kim2026optimizing} as the
target and a committee of Matlantis-PFP~v9.0.0~\citep{10.1038/s41467-022-30687-9,matlantis},
MACE-MPA-0~\citep{batatia2023foundation}, UMA-S-1.2~\citep{wood2026family}, and Orb-v3~\citep{rhodes2025orb}.
Seeded by the MLIP Arena homonuclear-diatomics family, the framework
found that SevenNet-Omni predicts an anomalous mid-range attraction for
the Cr$_2$, Mo$_2$, W$_2$, and Sb$_2$ dimers, consistently across all
three evaluated model channels: at $R = 4.5$~\AA{}, well beyond the
equilibrium bond length yet well inside the model's 6.0~\AA{} cutoff,
it still holds 17\,\% (Sb$_2$), 11\,\% (Mo$_2$), and 9\,\% (W$_2$) of
its own well depth, versus at most 4\,\% for the committee models
(Fig.~\ref{fig:omni_dimer_tail}); no committee model shows a
comparable attractive tail, and for Sb$_2$ Orb-v3 is instead
repulsive at mid-range, recorded as a separate side finding.
The Cr dimer is excluded from the quantitative comparison: it is a
notoriously difficult electronic-structure problem with an unusual
potential energy curve~\citep{purwanto2015auxiliary}, and the committee models
disagree even near equilibrium, leaving no label-free reference band
to judge the tail against.
The target model also binds more deeply near equilibrium, but the well
depths of the committee models themselves spread on a comparable
scale, so that deviation cannot be attributed label-free; the
finding is therefore reported on the tail, where no committee model
retains appreciable binding.

Although a careful visual inspection of the individual curves would
reveal the tail, this candidate anomaly is easily overlooked when models are
compared through the aggregate scores of the benchmark that seeded it.
The aggregate diatomics metrics of MLIP Arena are reference-free
statistics of each model's own curve, such as smoothness, tortuosity,
sign flips, and force--energy conservation, and do not compare the
predicted binding at a given distance against an external reference.

The escalation was also explicitly bounded.
The impact probes established where the candidate anomaly does \emph{not}
matter: when the same pairs are embedded in bulk, the excess pair
interaction is screened to within the committee spread, and NVE
energy drift is unaffected, confining the consequences to dilute
gas-phase thermodynamics such as dimerization equilibria and
nucleation onsets, where the 0.73~eV residual binding of Sb$_2$ at
4.5~\AA{} ($7\,kT$ at 1200~K) distorts pair-association weights by
orders of magnitude.
The escalation report records this domain boundary and justifies
escalation by the strength of the surrogate signal and the low
verification cost (32 two-atom single-point calculations), not by
breadth of impact: a finding less application-critical than those
in Section~\ref{sec:results}, and labeled as such before reaching the
human verifier.

The escalated report awaits human expert verification: while the verification card requires only 32 PBE single points, transition-metal dimers are notoriously difficult for single-reference DFT, so a definitive verdict on the true potential-energy surface would require costlier multi-reference calculations, which were deprioritized in favor of the more application-critical findings of Section~\ref{sec:results}.

\section{Limitations}
\label{app:limitations}
\paragraph{The search is bounded by the inspection specification.}
Hypothesis generation is anchored to the human-authored inspection
specification: the agents reliably probe the constraints and
diagnostics it documents and their neighborhoods, but they rarely
propose failure classes outside it. The quality of the search is
therefore capped by the quality of the inspection
specification, and extending the framework to
new kinds of failures currently requires a human to revise the
specification.

\paragraph{Human verification limits the number of findings.}
Every escalated candidate consumes expert attention and may require
new reference calculations, so only a small number of findings can be
verified per unit of expert time.
Extending agent authority into the verification stage may relax this bottleneck while keeping the final verdict with the human expert.

\paragraph{Framework outputs require manual polishing for
presentation.}
The escalation reports are designed for the verification decision,
not for publication: the figures and the report prose in this paper
were manually reworked from the framework's raw outputs. Automating
publication-quality reporting is left to future work.

\paragraph{The evaluation remains limited in scale.}
The present experiments are intended to establish feasibility and cover only a small number of search runs, target models, and failure modes. A more comprehensive evaluation should repeat the search across independent runs and broader tasks, define quantitative metrics such as verification yield, discovery cost, and run-to-run consistency, and compare the framework against random, rule-based, and expert-designed search strategies. Controlled ablations of the inspection specification, auxiliary committee, evidence memory, surrogate signals, and impact probes are also needed to determine which components contribute to successful failure mode discovery.

\end{document}